\documentclass{article}

\usepackage{microtype}
\usepackage{multicol}
\usepackage{multirow}
\usepackage{graphicx}
\usepackage{booktabs} 
\usepackage[normalem]{ulem}
\useunder{\uline}{\ul}{}
\usepackage{diagbox}

\usepackage{caption}
\usepackage{subcaption}

\usepackage{hyperref}

\usepackage[accepted]{icml2021}

\icmltitlerunning{Scaling Up Visual and Vision-Language Representation Learning With Noisy Text Supervision}

\begin{document}

\twocolumn[
\icmltitle{Scaling Up Visual and Vision-Language Representation Learning\\ With Noisy Text Supervision}



\icmlsetsymbol{equal}{*}

\begin{icmlauthorlist}
\icmlauthor{Chao Jia}{goo}
\icmlauthor{Yinfei Yang}{goo}
\icmlauthor{Ye Xia}{goo}
\icmlauthor{Yi-Ting Chen}{goo}
\icmlauthor{Zarana Parekh}{goo}
\icmlauthor{Hieu Pham}{goo}
\icmlauthor{Quoc V. Le}{goo}
\icmlauthor{Yunhsuan Sung}{goo}
\icmlauthor{Zhen Li}{goo}
\icmlauthor{Tom Duerig}{goo}
\end{icmlauthorlist}

\icmlaffiliation{goo}{Google Research}

\icmlcorrespondingauthor{Chao Jia}{chaojia@google.com}
\icmlcorrespondingauthor{Yinfei Yang}{yinfeiy@google.com}

\icmlkeywords{Visual Representation Learning, Visual Semantic Embedding, Multi-modal Retrieval}

\vskip 0.3in
]



\printAffiliationsAndNotice{}  

\begin{abstract}
Pre-trained representations are becoming crucial for many NLP and perception tasks. While representation learning in NLP has transitioned to training on raw text  without human annotations, visual and vision-language representations still rely heavily on curated training datasets that are 
expensive or require expert knowledge. For vision applications, representations are mostly learned using datasets with explicit class labels such as ImageNet or OpenImages.  For vision-language, popular datasets like Conceptual Captions, MSCOCO, or CLIP all involve a non-trivial data collection (and cleaning) process. This costly curation process limits the size of datasets and hence hinders the scaling of trained models. In this paper, we leverage a noisy dataset of over one billion image alt-text pairs, obtained without expensive filtering or post-processing steps in the Conceptual Captions dataset. A simple dual-encoder architecture learns to align visual and language representations of the image and text pairs using a contrastive loss. We show that the scale of our corpus can make up for its noise and leads to state-of-the-art representations even with such a simple learning scheme. Our visual representation achieves strong performance when transferred to classification tasks such as ImageNet and VTAB. The aligned visual and language representations enables zero-shot image classification and also set new state-of-the-art results on Flickr30K and MSCOCO image-text retrieval benchmarks, even when compared with more sophisticated cross-attention models. 
The  representations also enable cross-modality search with complex text and text + image queries.
\end{abstract}

\begin{figure*}[t]
    \centering
    \includegraphics[scale=0.4]{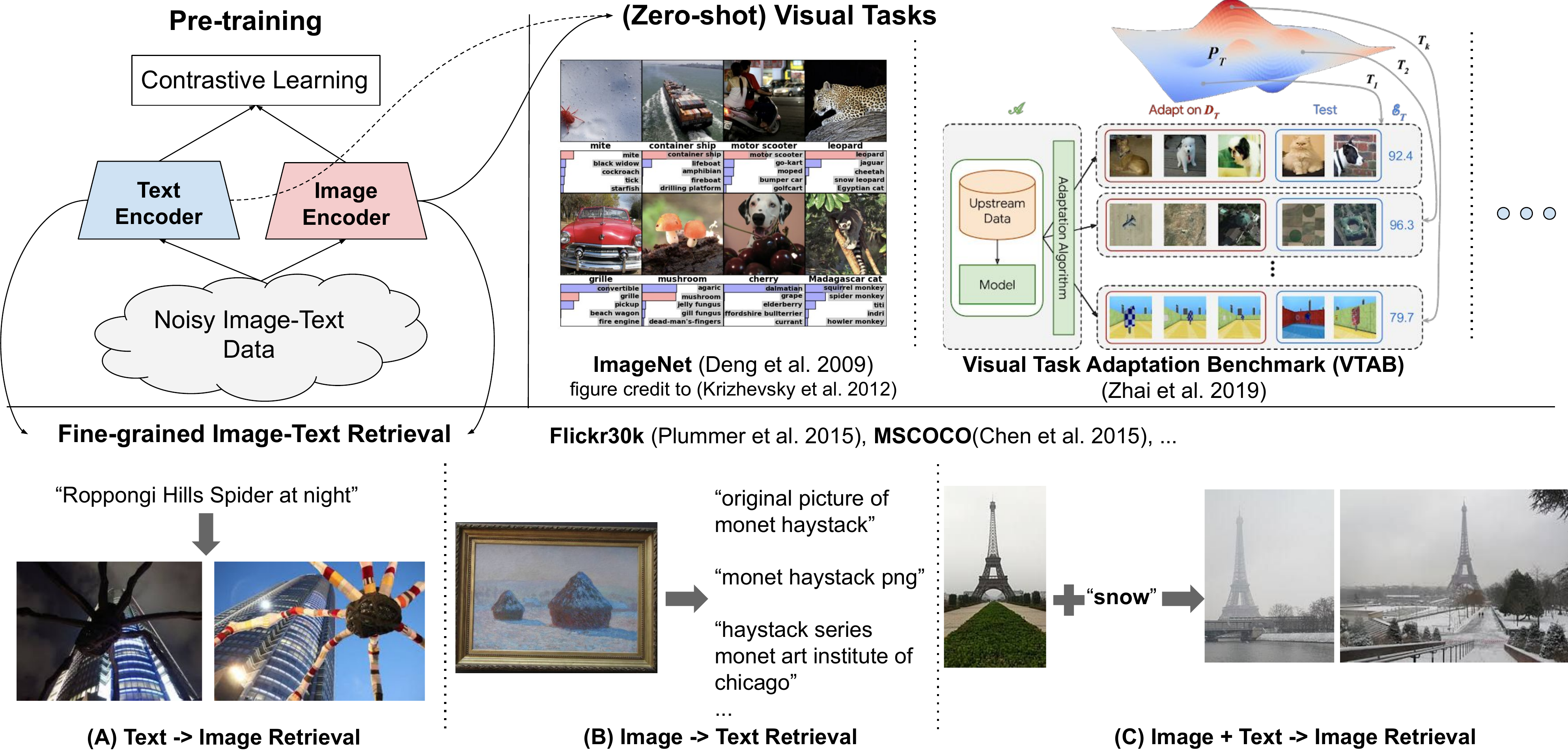}
    \vskip -0.1in
    \caption{A summary of our method, ALIGN. Visual and language representations are jointly learned from noisy image alt-text data. The representations can be used for vision-only or vision-language task transfer. Without any fine-tuning, ALIGN powers zero-shot visual classification and cross-modal search including image-to-text search, text-to-image search and even search with joint image+text queries.}
    \label{fig:diagram}
    \vskip -0.1in
\end{figure*}

\section{Introduction}
In the existing literature, visual 
and vision-language representation learning are mostly studied separately with different training data sources. In the vision domain, 
pre-training
on large-scale supervised data such as ImageNet \cite{deng:imagenet}, OpenImages \cite{kuznetsova:openimages}, and JFT-300M \cite{sun:jft, kolensnikov:bit} has proven to be critical for improving performance on downstream tasks via transfer learning. Curation of such pre-training datasets 
requires heavy work on data gathering, sampling, and human annotation, and hence is difficult to scale.\looseness=-1

Pre-training has also become the de-facto approach in vision-language modeling \cite{lu:vilbert, chen:uniter, li:oscar}. However, vision-language pre-training datasets such as Conceptual Captions \cite{sharma:cc3m}, Visual Genome Dense Captions \cite{krishna:visualgenome}, 
and ImageBERT
\cite{qi:imagebert} require even heavier work on human annotation, semantic parsing, 
cleaning and balancing. As a result, the scales of these datasets are only in the realm of $\sim$10M examples. This is at least an order of magnitude smaller than their counterparts in the vision domain, and much smaller than large corpora of text from the internet for 
NLP
pre-training (e.g., \citet{devlin:bert, radford:gpt2, yang:xlnet, liu:roberta, raffel:t5}).\looseness=-1

In this work, 
we leverage a dataset of over one billion noisy image alt-text pairs to scale visual and vision-language representation learning. 
We follow the procedures described in the Conceptual Captions dataset~\cite{sharma:cc3m} to have a large noisy dataset. But instead of applying the complex filtering and post-processing steps as proposed by~\cite{sharma:cc3m} to clean the dataset, we only apply simple frequency-based filtering. 
The resulting dataset is noisy, but is two orders of magnitude larger than the Conceptual Captions dataset. We show that visual and vision-language representations pre-trained on our exascale dataset achieve very strong performance on a wide range of tasks. 

To train our model, we use an objective that aligns the visual and language representations in a shared latent embedding space using a simple dual-encoder architecture. Similar objectives has been applied to learning visual-semantic embeddings (VSE) \cite{frome:devise, faghri:vse++}. We name our model \textbf{ALIGN}: \textbf{A} \textbf{L}arge-scale \textbf{I}ma\textbf{G}e and \textbf{N}oisy-text embedding. Image and text encoders are learned via a contrastive loss (formulated as normalized softmax) that pushes the embeddings of matched image-text pair together while pushing those of non-matched image-text pair apart. This is one of the most effective loss functions for both self-supervised \cite{chen:simclr} and supervised \cite{zhai:norm_softmax, musgrave:metric_check} representation learning. Considering paired texts as fine-grained labels of images, our image-to-text contrastive loss is analogous to the conventional label-based classification objective; and the key difference is that the text encoder generates the ``label" weights. The top-left of Figure \ref{fig:diagram} summarizes the method we use in ALIGN.\looseness-1

The aligned image and text representations are naturally suited for cross-modality matching/retrieval tasks and achieve state-of-the-art (SOTA) results in corresponding benchmarks. For instance, ALIGN outperforms the previous SOTA method by over 7\% in most zero-shot and fine-tuned R@1 metrics in Flickr30K and MSCOCO. Moreover, such cross-modality matching naturally enables zero-shot image classification when feeding the classnames into the text encoder, achieving 76.4\% top-1 accuracy in ImageNet without using any of its training samples. The image representation itself also achieves superior performance in various downstream visual tasks. For example, ALIGN achieves 88.64\% top-1 accuracy in ImageNet.  Figure \ref{fig:diagram}-bottom shows the cross-modal retrieval examples that come from a real retrieval system built by ALIGN.

\vspace{-2mm}
\section{Related Work}

High-quality visual representations for classification or retrieval are usually pre-trained on large-scale labeled datasets \cite{mahajan:wsl, kolensnikov:bit, dosovitskiy:vit, juan:graphrise}. Recently, self-supervised \citep{chen:simclr, tian:cmc, he:moco, misra:pirl, li:pcl, grill:byol, caron:swav} and semi-supervised learning \citep{yalniz2019billion, xie:noisy_student, pham:mpl} have been studied as alternative paradigms. 
However, models trained by these methods so far show limited transferability to downstream tasks~\cite{zoph2020rethinking}.\looseness=-1 


Leveraging images and natural language captions is another direction of learning visual representations.~\citet{joulin:flickr,li:ngram,desai:virtex,sariyildiz2020learning,zhang2020contrastive} show that a good visual representation can be learned by predicting the captions from images, which inspires our work. These works are however limited to small datasets such as Flickr~\cite{joulin:flickr,li:ngram} and COCO Captions~\cite{desai:virtex,sariyildiz2020learning}, and the resulting models don't produce a vision-language representation that is needed for tasks like cross-modal retrieval.



In the vision-language representation learning domain, visual-semantic embeddings (VSE)~\citep{frome:devise,faghri:vse++} and improved versions (e.g., leveraging object detectors, dense feature maps, or multi-attention layers)~\citep{socher-etal-2014-grounded,karpathy:2014,kiros:2014,nam:dan,li:vsrn,teran,chen:vsepooling} have been proposed.
Recently more advanced models emerge with cross-modal attention layers \cite{liu:mia, lu:vilbert, chen:uniter, pixel-bert} and show superior performance in image-text matching tasks. However, they are orders of magnitudes slower and hence impractical for image-text retrieval systems in the real world. In contrast, our model inherits the simplest VSE form, but still outperforms all previous cross-attention models in image-text matching benchmarks.\looseness=-1

Closely related to our work is CLIP \cite{radford:clip}, which proposes visual representation learning via natural language supervision in a similar contrastive learning setting. Besides using different vision and language encoder architectures, the key difference is on training data: ALIGN follows the natural distribution of image-text pairs from the raw alt-text data, while CLIP collects the dataset by first constructing an allowlist of high-frequency visual concepts from English Wikipedia. We demonstrate that strong visual and vision-language representations can be learned with a dataset that doesn't require expert knowledge to curate.\looseness-1

\vspace{-2mm}
\section{A Large-Scale Noisy Image-Text Dataset} \label{data}

\begin{figure}[!htb]
\begin{center}
    \centerline{\includegraphics[width=\linewidth]{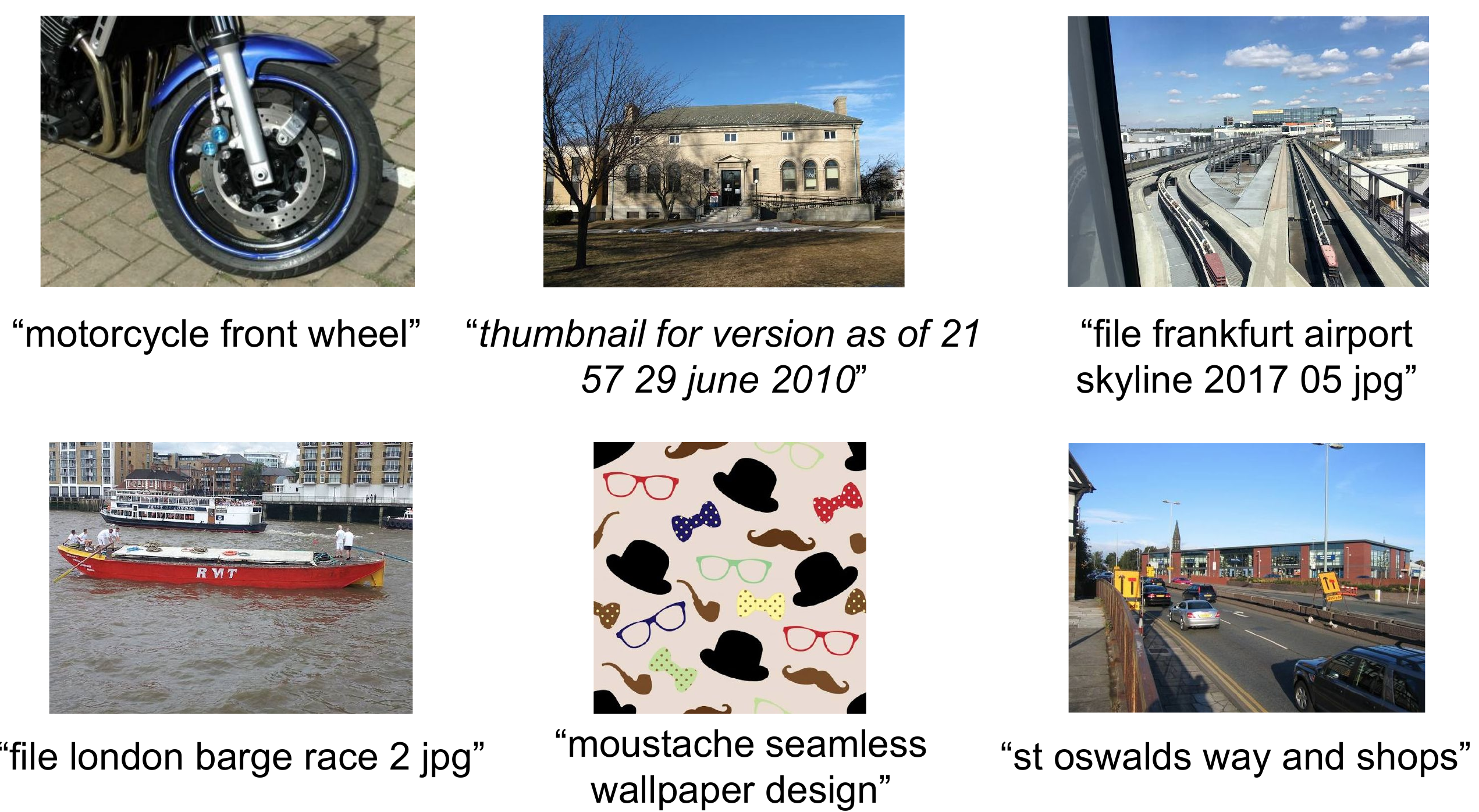}}
    \vskip -0.1in
    \caption{Example image-text pairs randomly sampled from the training dataset of ALIGN. One clearly noisy text annotation is marked in \textit{italics}.}
    \label{fig:examples}
    \vskip -0.2in
\end{center}
\end{figure}

The focus of our work is to scale up visual and vision-language representation learning. For this purpose, we resort to a much larger dataset than existing ones. Specifically, we follow the methodology of constructing Conceptual Captions dataset~\cite{sharma:cc3m} to get a version of raw English alt-text data (image and alt-text pairs). The Conceptual Captions dataset was cleaned by heavy filtering and post-processing. Here, for the purpose of scaling, we trade quality for scale by relaxing most of the cleaning steps in the original work. Instead, we only apply minimal frequency-based filtering as detailed below. The result is a much larger (1.8B image-text pairs) but noisier dataset. 
Figure \ref{fig:examples} shows some sample image-text pairs from the dataset.\looseness-1

\vspace{-2mm}
\paragraph{Image-based filtering.} Following~\citet{sharma:cc3m}, we remove pornographic images and keep only images whose shorter dimension is larger than 200 pixels and aspect ratio is smaller than 3. Images with more than 1000 associated alt-texts are discarded. To ensure that we don't train on test images, we also remove duplicates or near-duplicates of test images in all downstream evaluation datasets (e.g., ILSVRC-2012, Flickr30K, and MSCOCO). See Appendix \ref{sec:near-dup} for more details.

\paragraph{Text-based filtering.} We exclude alt-texts that are shared by more than 10 images. These alt-texts are often irrelevant to the content of the images (e.g., ``1920x1080", ``alt\textunderscore img", and ``cristina"). We also discard alt-texts that contain any rare token (outside of 100 million most frequent unigrams and bigrams from the raw dataset), and those that are either too short ($<$3 unigrams) or too long ($>$20 unigrams). This removes noisy texts like ``image\_tid 25\&id mggqpuweqdpd\&cache 0\&lan\_code 0", or texts that are too generic to be useful.
\vspace{-2mm}

\section{Pre-training and Task Transfer}
\subsection{Pre-training on Noisy Image-Text Pairs}
We pre-train ALIGN using a dual-encoder architecture. The model consists of a pair of image and text encoders with a cosine-similarity combination function at the top. We use EfficientNet with global pooling (without training the 1x1 conv layer in the classification head) as the image encoder and BERT with [CLS] token embedding as the text embedding encoder (we generate 100k wordpiece vocabulary from our training dataset). A fully-connected layer with linear activation is added on top of BERT encoder to match the dimension from the image tower. Both image and text encoders are trained from scratch. 

The image and text encoders are optimized via normalized softmax loss \cite{zhai:norm_softmax}. In training, we treat matched image-text pairs as positive and all other random image-text pairs that can be formed in a training batch as negative.\looseness=-1

We minimize the sum of two losses: one for image-to-text classification
\vspace{-0.2cm}
\begin{equation}\label{eq:i2t_loss}
\small
    L_{i2t} = -\frac{1}{N}\sum_i^N\log{\frac{\exp(x_i^\top y_i / \sigma)}{\sum_{j=1}^{N} \exp(x_i^\top y_j / \sigma)}}
\end{equation}
\vspace{-0.2cm}
and the other for text-to-image classification
\begin{equation}\label{eq:t2i_loss}
\small
    L_{t2i} = -\frac{1}{N}\sum_i^N\log{\frac{\exp(y_i^\top x_i / \sigma)}{\sum_{j=1}^{N} \exp(y_i^\top x_j / \sigma)}}
\end{equation}
\vspace{-0.4cm}

Here, $x_i$ and $y_j$ are the normalized embedding of image in the $i$-th pair and that of text in the $j$-th pair, respectively. $N$ is the  batch size, and $\sigma$ is the temperature to scale the logits. For in-batch negatives to be more effective, we concatenate embeddings from all computing cores to form a much larger batch. The temperature variable is crucial as both image and text embeddings are L2-normalized. Instead of manually sweeping for the optimal temperature value, we find that it can be effectively learned together with all the other parameters.

\vspace{-0.2cm}
\subsection{Transferring to Image-Text Matching \& Retrieval}

We evaluate ALIGN models on image-to-text and text-to-image retrieval tasks, with and without finetuning. Two benchmark datasets are considered: Flickr30K \cite{plummer:flickr30k} and MSCOCO \cite{chen:coco}. 
We also evaluate ALIGN on Crisscrossed Captions (CxC)~\cite{parekh:cxc}, which is an extension of MSCOCO with additional human semantic similarity judgments for caption-caption, image-image, and image-caption pairs.
With extended annotations, CxC enables four intra- and inter-modal retrieval tasks including image-to-text, text-to-image, text-to-text, and image-to-image retrieval, and three semantic similarity tasks including semantic textual similarity (STS), semantic image similarity (SIS), and semantic image-text similarity (SITS). 
As the training set is identical to the original MSCOCO, we can directly evaluate the MSCOCO fine-tuned ALIGN model on CxC annotations.\looseness=-1
\vspace{-0.2cm}
\subsection{Transferring to Visual Classification}

We first apply zero-shot transfer of ALIGN to visual classification tasks on ImageNet ILSVRC-2012 benchmark \cite{deng:imagenet} and its variants including ImageNet-R(endition) \cite{imagenet-r} (non-natural images such as art, cartoons, sketches), ImageNet-A(dversarial) \cite{imagenet-a} (more challenging images for ML models), and ImageNet-V2 \cite{imagenet-v2}. All of these variants follow the same set (or a subset) of ImageNet classes, while the images in ImageNet-R and ImageNet-A are sampled from drastically different distributions from ImageNet.

We also transfer the image encoder to downstream visual classification tasks. For this purpose, we use the  ImageNet as well as a handful of smaller fine-grained classification datasets such as Oxford Flowers-102 \cite{nilsback:flowers}, Oxford-IIIT Pets \cite{parkhi:pets}, Stanford Cars \cite{krause:cars196}, and Food101 \cite{bossard:food101}. For ImageNet, results from two settings are reported: training the top classification layer only (with frozen ALIGN image encoder) and fully fine-tuned. Only the latter setting is reported for fine-grained classification benchmarks. Following \citet{kolensnikov:bit}, we also evaluate the robustness of our model on Visual Task Adaptation Benchmark (VTAB) \cite{zhai:vtab} which consists of 19 diverse (covering subgroups of natural, specialized and structured image classification tasks) visual classification tasks with 1000 training samples each. 
\looseness-1
\vspace{-0.3cm}
\section{Experiments and Results}\label{sec:experiment_result}

We train our ALIGN models from scratch, using the open-sourced implementation of EfficientNet as the image encoder and BERT as the text encoder. Unless in the ablation study, 
we use the results of ALIGN where the image encoder is EfficientNet-L2 and the text encoder is BERT-Large.
The image encoder is trained at resolution of 289 $\times$ 289 pixels no matter what EfficientNet variant is used. We first resize input images to 346 $\times$ 346 resolution and then perform random crop (with additional random horizontal flip) in training and central crop in evaluation. For BERT we use wordpiece sequence of maximum 64 tokens since the input texts are no longer than 20 unigrams. The softmax temperature variable is initialized as 1.0 (this temperature variable is shared between image-to-text loss and text-to-image loss) and we use 0.1 as label smoothing parameter in the softmax losses. We use LAMB optimizer \cite{you:lamb}\footnote{We tried SGD with momentum and ADAM which are known to work well for CNNs and BERT respectively. LAMB appears to be a better choice for training both image and text encoders.} with weight decay ratio 1e-5. The learning rate is warmed up linearly to 1e-3 from zero in 10k steps, and then linearly decay to zero in 1.2M steps ($\sim$12 epochs). We train the model on 1024 Cloud TPUv3 cores with 16 positive pairs on each core. Therefore the total effective batch size is 16384.\looseness=-1

\begin{table*}[h!]
\begin{center}
\caption{Image-text retrieval results on Flickr30K and MSCOCO datasets (zero-shot and fine-tuned). ALIGN is compared with ImageBERT~\cite{qi:imagebert}, UNITER~\cite{chen:uniter}, CLIP~\cite{radford:clip}, GPO~\cite{chen:vsepooling}, ERNIE-ViL~\cite{yu:ernie-vil}, VILLA~\cite{gan:villa}, and Oscar~\cite{li:oscar}.}
\label{tab:flickr30k_mscoco_result}
\vspace{2mm}
\begin{small}
\resizebox{0.98\linewidth}{!}{ %
\begin{tabular}{ll|rrrrrr|rrrrrr}
\toprule
& & \multicolumn{6}{c}{Flickr30K (1K test set)} & \multicolumn{6}{c}{MSCOCO (5K test set)} \\
& & \multicolumn{3}{c}{image $\to$ text} & \multicolumn{3}{c}{text $\to$ image} & \multicolumn{3}{c}{image $\to$ text} & \multicolumn{3}{c}{text $\to$ image} \\
\midrule
& & R@1 & R@5 & R@10 & R@1 & R@5 & R@10 & R@1 & R@5 & R@10 & R@1 & R@5 & R@10 \\
\multirow{4}{*}{Zero-shot} & ImageBERT & 70.7 & 90.2 & 94.0 & 54.3 & 79.6 & 87.5 & 44.0 & 71.2 &  80.4 & 32.3 & 59.0 & 70.2 \\
& UNITER & 83.6 & 95.7 & 97.7 & 68.7 & 89.2 & 93.9 & - & - & - & - & - & - \\
& CLIP & 88.0 & 98.7 & 99.4 & 68.7 & 90.6 & 95.2 & 58.4 & 81.5 & 88.1 & 37.8 & 62.4 & 72.2 \\
& \textbf{ALIGN} & \textbf{88.6} & \textbf{98.7} & \textbf{99.7} & \textbf{75.7} & \textbf{93.8} & \textbf{96.8} & \textbf{58.6} & \textbf{83.0} & \textbf{89.7} & \textbf{45.6} & \textbf{69.8} & \textbf{78.6} \\
\hline
\multirow{6}{*}{Fine-tuned} & GPO & 88.7 & 98.9 & 99.8 & 76.1 & 94.5 & 97.1 & 68.1 & 90.2 & - & 52.7 & 80.2 & - \\
& UNITER & 87.3 & 98.0 & 99.2 & 75.6 & 94.1 & 96.8 & 65.7 & 88.6 & 93.8 & 52.9 & 79.9 & 88.0 \\
& ERNIE-ViL & 88.1 & 98.0 & 99.2 & 76.7 & 93.6 & 96.4 & - & - & - & - & - & - \\
& VILLA & 87.9 & 97.5 & 98.8 & 76.3 & 94.2 & 96.8 & - & - & - & - & - & - \\
& Oscar & - & - & - & - & - & - & 73.5 & 92.2 & 96.0 & 57.5 & 82.8 & \textbf{89.8} \\
& \textbf{ALIGN} & \textbf{95.3} & \textbf{99.8} & \textbf{100.0} & \textbf{84.9} & \textbf{97.4} & \textbf{98.6} & \textbf{77.0} & \textbf{93.5} & \textbf{96.9} & \textbf{59.9} & \textbf{83.3} & \textbf{89.8} \\
\bottomrule
\end{tabular}
} %
\end{small}
\end{center}
\vskip -0.1in
\end{table*}

\begin{table*}[t]
\begin{center}
\caption{Multimodal retrieval performance on Crisscrossed Captions~(CxC) dataset. ALIGN is compared with VSE++~\cite{faghri:vse++}, VSRN~\cite{li:vsrn}, DE\textsubscript{I2T}~\cite{parekh:cxc}, and DE\textsubscript{T2T+I2T}~\cite{parekh:cxc}.}
\label{tab:cxc_results}
\vspace{2mm}
\begin{small}
\begin{tabular}{l|rrr|rrr|rrr|rrr}
\toprule
& \multicolumn{3}{c|}{image $\to$ text} & \multicolumn{3}{c|}{text $\to$ image} & \multicolumn{3}{c|}{text $\to$ text} & \multicolumn{3}{c}{image $\to$ image} \\
\midrule
& R@1 & R@5 & R@10 & R@1 & R@5 & R@10 & R@1 & R@5 & R@10 & R@1 & R@5 & R@10 \\
VSE++                     & 43.1 & 74.3 & 84.2 & 32.5 & 62.7 & 75.4 & 38.7 & 62.3 & 72.2 & 36.4 & 70.4 & 81.3 \\
VSRN                      & 52.4 & 81.9 & 90.0 & 40.1 & 71.1 & 81.5 & 41.0 & 64.8 & 74.5 & 44.2 & 76.7 & 86.2 \\ 
DE\textsubscript{I2T}     & 53.9 & 82.7 & 91.2 & 39.8 & 70.2 & 80.9 & 26.0 & 47.1 & 57.5 & 38.3 & 74.1 & 85.0 \\
DE\textsubscript{T2T+I2T} & 55.9 & 84.2 & 91.8 & 41.7 & 72.3 & 83.0 & 42.4 & 64.9 & 74.0 & 38.5 & 73.6 & 84.9 \\
\textbf{ALIGN}            & \textbf{78.1} & \textbf{94.3} & \textbf{97.4} & \textbf{61.8} & \textbf{84.9} & \textbf{91.1} & \textbf{45.4} & \textbf{66.8} & \textbf{75.2} & \textbf{49.4} & \textbf{81.4} & \textbf{89.1} \\
\bottomrule
\end{tabular}
\end{small}
\end{center}
\vspace{-5mm}
\end{table*}

\begin{table}[t]
\centering
\small
\caption{Spearman's R Bootstrap Correlation ($\times 100$) on Crisscrossed Captions~(CxC) dataset. ALIGN is compared with VSE++~\cite{faghri:vse++}, VSRN~\cite{li:vsrn}, DE\textsubscript{I2T}~\cite{parekh:cxc}, and DE\textsubscript{T2T+I2T}~\cite{parekh:cxc}.}
\label{tab:cxc_spearman_corr}
\vspace{2mm}
\resizebox{0.99\linewidth}{!}{ %
\begin{tabular}{l|rrrr}
\toprule
\multirow{2}{*}{\textbf{Model}} & \multicolumn{1}{c}{\textbf{STS}} & \multicolumn{1}{c}{\textbf{SIS}} & \multicolumn{1}{c}{\textbf{SITS}} & \multicolumn{1}{c}{\textbf{Mean Avg}} \\
& avg $\pm$ std & avg $\pm$ std & avg $\pm$ std & \\
\hline
VSE++                      & \textbf{74.4$\pm$0.4} & 73.3$\pm$0.9 & 55.2$\pm$1.5 & 67.6 \\
VSRN                       & 73.0$\pm$0.4 & 70.1$\pm$1.0 & 60.4$\pm$1.3 & 67.8 \\
DE\textsubscript{I2T}      & 50.9$\pm$0.6 & \textbf{81.3$\pm$0.7} & 61.6$\pm$1.4 & 64.6 \\
DE\textsubscript{T2T+I2T}  & 74.2$\pm$0.4 & 74.5$\pm$0.9 & 61.9$\pm$1.3 & 70.2 \\
\textbf{ALIGN}             & 72.9$\pm$0.4 & 77.2$\pm$0.8 & \textbf{67.6$\pm$1.2} & \textbf{72.6}\\
\bottomrule
\end{tabular}
}
\vspace{-5mm}
\end{table}

\vspace{-2mm}
\subsection{Image-Text Matching \& Retrieval}

We evaluate ALIGN on Flickr30K and MSCOCO cross-modal retrieval benchmarks, in both zero-shot and fully fine-tuned settings. We follow \cite{karpathy:image-text2015} and most existing works to obtain the train/test splits. Specifically, for Flickr30K, we evaluate on the standard 1K test set, and finetune on the 30k training set. For MSCOCO, we evaluate on the 5K test set, and finetune on 82K training plus 30K additional validation images that are not in the 5K validation or 5K test sets.

During fine-tuning, the same loss function is used. But there can be false negatives when the batch size is comparable to the total number of training samples. So we reduce the global batch size from 16384 to 2048. We also reduce the initial learning rate to 1e-5 and train for 3K and 6K steps (with linear decay) respectively on Flickr30K and MSCOCO. All the other hyper-parameters are kept the same as pre-training.\looseness=-1

Table \ref{tab:flickr30k_mscoco_result} shows that, compared to previous works, ALIGN achieves SOTA results in all metrics of Flickr30K and MSCOCO benchmarks. In the zero-shot setting, ALIGN gets more than 7\% improvement in image retrieval task compared to the previous SOTA, CLIP~\cite{radford:clip}. With fine-tuning, ALIGN outperforms all existing methods by a large margin, including those that employ more complex cross-modal attention layers such as ImageBERT~\cite{qi:imagebert}, UNITER~\cite{chen:uniter}, ERNIE-ViL~\cite{yu:ernie-vil}, VILLA~\cite{gan:villa} and Oscar~\cite{li:oscar}.\looseness=-1

Table \ref{tab:cxc_results} reports the performance of ALIGN on Crisscrossed Captions (CxC) retrieval tasks. Again, ALIGN achieves  SOTA results in all metrics, especially by a large margin on image-to-text (+22.2\% R@1) and text-to-image (20.1\% R@1) tasks.
Table \ref{tab:cxc_spearman_corr} shows that ALIGN also outperforms the previous SOTA on SITS task with an improvement of 5.7\%.
One interesting observation is that, despite being much better on inter-modal tasks, ALIGN is not as impressive on intra-modal tasks.
For instance, the improvements on text-to-text and image-to-image retrieval tasks (in particular the former) are less significant compared to those on image-to-text and text-to-image tasks.
The performance on STS and SIS tasks is also slightly worse than VSE++ and DE\textsubscript{I2T}. 
We suspect it is because the training objective of ALIGN focuses on cross-modal (image-text) matching instead of intra-modal matching.
\citet{parekh:cxc} suggest multitask learning could produce more balanced representations.
We leave it to the future work.

\vspace{-2mm}

\subsection{Zero-shot Visual Classification}

If we directly feed the texts of classnames into the text encoder, ALIGN is able to classify images into candidate classes via image-text retrieval. Table \ref{tab:zs_imagenet_result} compares ALIGN with CLIP on Imagenet and its variants. Similar to CLIP, ALIGN shows great robustness on classification tasks with different image distributions. In order to make a fair comparison, we use the same prompt ensembling method as CLIP. Each classname is expanded with a set of prompt templates defined by CLIP such as ``A photo of a \{classname\}''. The class embedding is computed by averaging the embeddings of all templates followed by an L2-normalization. We find that such ensembling gives 2.9\% improvement on ImageNet top-1 accuracy.\looseness=-1

\begin{table}[h!]
\vspace{-3mm}
\begin{center}
\caption{Top-1 Accuracy of zero-shot transfer of ALIGN to image classification on ImageNet and its variants.}
\label{tab:zs_imagenet_result}
\vskip -0.1in
\begin{small}
\resizebox{\linewidth}{!}{ %
\begin{tabular}{l|llll}
\toprule
Model & ImageNet & ImageNet-R & ImageNet-A & ImageNet-V2 \\
\midrule
CLIP  & 76.2 & 88.9 & \textbf{77.2} & \textbf{70.1} \\
\textbf{ALIGN} & \textbf{76.4} & \textbf{92.2} & 75.8 & \textbf{70.1} \\
\bottomrule
\end{tabular}
}
\end{small}
\end{center}
\vspace{-3mm}
\end{table}

\begin{table*}[t!]
\vspace{-3mm}
\begin{center}
\caption{ImageNet classification results. ALIGN is compared with WSL~\cite{mahajan:wsl}, CLIP~\cite{radford:clip}, BiT~\cite{kolensnikov:bit}, ViT~\cite{dosovitskiy:vit}, NoisyStudent~\cite{xie:noisy_student}, and Meta-Pseudo-Labels~\cite{pham:mpl}.}
\label{tab:imagenet_result}
\begin{small}
\begin{tabular}{l|llll}
\toprule
Model (backbone) & Acc@1 w/ frozen features & Acc@1 & Acc@5\\
\midrule
WSL (ResNeXt-101 32x48d) & 83.6 & 85.4 & 97.6 \\
CLIP (ViT-L/14) & 85.4 & - & - \\
BiT (ResNet152~x~4) & - & 87.54 & 98.46 \\
NoisyStudent (EfficientNet-L2) & - & 88.4 & 98.7 \\
ViT (ViT-H/14) & - & 88.55 & - \\
Meta-Pseudo-Labels (EfficientNet-L2) & - & \textbf{90.2} & \textbf{98.8} \\
\textbf{ALIGN} (EfficientNet-L2) & \textbf{85.5} & 88.64 & 98.67 \\ 
\bottomrule
\end{tabular}
\end{small}
\end{center}
\vspace{-5mm}
\end{table*}

\subsection{Visual Classification w/ Image Encoder Only}

On the ImageNet benchmark, we first freeze the learned visual features and only train the classification head. Afterwards we fine-tune all layers. We use basic data augmentations including random cropping (same as in \citet{szegedy:inception}) and horizontal flip. In evaluation we apply a single central crop with ratio of 0.875. Following \citet{touvron:fixres}, we use 0.8 scale ratio between training and evaluation to mitigate the resolution discrepancy introduced by random crop. Specifically, train/eval resolution is 289/360 with frozen visual features, and is 475/600 when fine-tuning all variables.\looseness=-1

In both stages of training, we use a global batch size of 1024, SGD optimizer with momentum 0.9, and learning rate decayed every 30 epochs with ratio 0.2 (100 epochs in total). Weight decay is set to zero. With frozen visual features, we use the initial learning rate of 0.1. When fine-tuning all layers with use the initial learning rate of 0.01, and use 10x smaller learning rate on the  backbone network compared to the classification head.\looseness-1

Table \ref{tab:imagenet_result} compares ALIGN with previous methods on the ImageNet benchmark. With frozen features, ALIGN slightly outperforms CLIP and achieves SOTA result of 85.5\% top-1 accuracy. After fine-tuning ALIGN achieves higher accuracy than BiT and ViT models, and is only worse than Meta Pseudo Labels which requires deeper interaction between ImageNet training and large-scale unlabeled data. Compared to NoisyStudent and Meta-Pseudeo-Labels which also use EfficientNet-L2, ALIGN saves 44\% FLOPS by using smaller test resolution (600 instead of 800).

In VTAB eval, we follow a hyper-parameter sweep as shown in the Appendix I in \cite{zhai:vtab} with 50 trials for each task. Each task is trained on 800 images and the hyperparameters are selected using the validation set of 200 images. After the sweep, the selected hyperparameters are used to train on the combined training and validation splits of 1000 images for each task. Table \ref{tab:vtab_result} reports the mean accuracy (including the breakdown results on each subgroup) with standard deviation from three fine-tuning runs and shows that ALIGN outperforms BiT-L \cite{kolensnikov:bit} with similar hyper-parameter selection method applied.

\begin{table}[h!]
\vspace{-5mm}
\begin{center}
\caption{VTAB (19 tasks) comparison between ALIGN and BiT-L.}
\label{tab:vtab_result}
\vskip -0.1in
\label{tab:vtab}
\begin{small}
\resizebox{\linewidth}{!}{ %
\begin{tabular}{l|llll}
\toprule
Model & All tasks & Natural & Specialized & Structured \\
\midrule
Bit-L  & 78.72 & - & - & - \\
\textbf{ALIGN} & \textbf{79.99$\pm$0.15} & 83.38 & 87.56 & 73.25 \\
\bottomrule
\end{tabular}
}
\end{small}
\end{center}
\vspace{-5mm}
\end{table}

To evaluate on smaller fine-grained classification benchmarks, we adopt a simple fine-tuning strategy for all tasks. We use the same data augmentation and optimizer as in ImageNet fine-tuning. Similarly, we first train the classification head and then fine-tune all layers, except with batch norm statistics frozen. The train/eval resolution is fixed at 289/360. We use batch size 256 and weight decay 1e-5. The initial learning rate is set to 1e-2 and 1e-3 respectively, with cosine learning rate decay in 20k steps. Table \ref{tab:small_result} compares ALIGN with BiT-L~\cite{kolensnikov:bit} and SAM \cite{foret:sam} which both apply same fine-tuning hyper-parameters for all tasks.\footnote{ViT \cite{dosovitskiy:vit} uses different hyper-parameters for different tasks and hence is not included in comparison.} For small tasks like these, details in fine-tuning matter. So we list the baseline results in \cite{foret:sam} without using SAM optimization for a fairer comparison. Our result (average of three runs) is comparable to the SOTA results without tweaking on optimization algorithms.

\begin{table}[h!]
\vspace{-5mm}
\begin{center}
\caption{Transfer learning results on Fine-grained Classification Tasks. BiT-L~\cite{kolensnikov:bit} was trained with ResNet152~x~4 whereas SAM-baseline, SAM-final~\cite{foret:sam} and ALIGN were trained with EfficientNet-L2.}
\label{tab:small_result}
\vskip 0.05in
\begin{small}
\begin{tabular}{l|ccccc}
\toprule
\multirow{2}{*}{Model} & Oxford  & Oxford & Stanford & \multirow{2}{*}{Food101} \\
& Flowers & Pets & Cars & \\
\midrule
BiT-L& 99.63 & 96.62 & -  & - \\
SAM-baseline
& 99.60 & 96.92 & 95.07 & 96.03 \\
SAM-final
& \textbf{99.65} & \textbf{97.10} & 95.96  & \textbf{96.18} \\
\bf{ALIGN}
& \textbf{99.65} & 96.19 & \textbf{96.13} & 95.88 \\
\bottomrule
\end{tabular}
\end{small}
\end{center}
\vspace{-5mm}
\end{table}

\section{Ablation Study}\label{sec:ablation}

In the ablation study, we compare model performance mostly on MSCOCO zero-shot retrieval and ImageNet K-Nearest-neighbor (KNN) tasks.\footnote{For each image in the validation set of ImageNet, we retrieve its nearest neighbors from the training set w/ pre-trained image encoder. Recall@K metric is calculated based on if the groundtruth label of the query image appears in the top-K retrieved images.} We find these two metrics are representative and correlate well with other metrics reported in the section above. If not mentioned, hyper-parameters other than the ablated factor are kept the same as in the baseline model.

\subsection{Model Architectures}
We first study the performance of ALIGN models using different image and text backbones. We train EfficientNet from B1 to L2 for the image encoder and BERT-Mini to BERT-Large for the text encoder. We add an additional fully-connected layer with linear activation on top of B1, B3, B5 and L2 globally-pooled features to match the output dimension of B7 (640). A similar linear layer is added to all text encoders. We reduce the training steps to 1M in ablation to save some runtime.

\begin{figure*}[t]
\begin{center}
    \centerline{\includegraphics[width=0.88\linewidth]{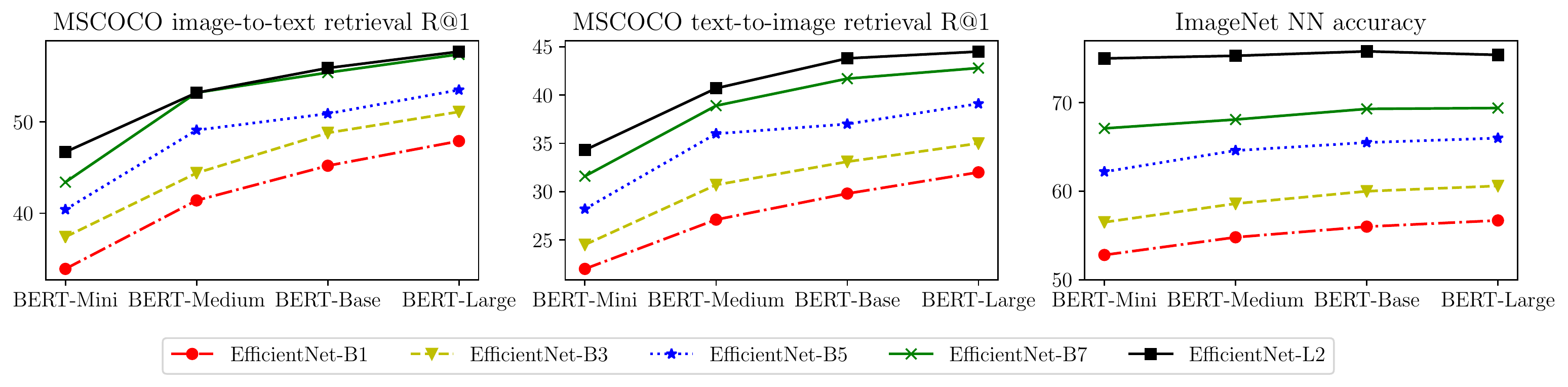}}
    \vskip -0.15in
    \caption{Zero-shot image-text retrieval and ImageNet KNN accuracy@1 with different image and text encoder sizes.}
    \label{fig:ablation_capacity}
    \vskip -0.3in
\end{center}
\end{figure*}

Figures \ref{fig:ablation_capacity} shows MSCOCO zero-shot retrieval and ImageNet KNN results with different combinations of image and text backbones. Model quality improves nicely with larger backbones except that the ImageNet KNN metric starts to saturate from BERT-Base to BERT-Large with EfficientNet-B7 and EfficientNet-L2. As expected, scaling up image encoder capacity is more important for vision tasks (e.g., even with BERT-Mini text tower, L2 performs better than B7 with BERT-Large). In image-text retrieval tasks the image and text encoder capacities are equally important. Based on the nice scaling property shown in Figure \ref{fig:ablation_capacity}, we only fine-tune the model with EfficientNet-L2 + BERT-Large as reported in Section \ref{sec:experiment_result}. \looseness=-1


We then study key architecture hyperparameters including embedding dimensions,
 number of random negatives in the batch, and the softmax temperature. Table \ref{tab:ablation_params} compares a number of model variants to a baseline model (first row) trained with the following settings: EfficientNet-B5 image encoder, BERT-Base text encoder, embedding dimension 640, all negatives in the batch, and a learnable softmax temperature.

Rows 2-4 of Table~\ref{tab:ablation_params} show that model performance improves with higher embedding dimensions. Hence, we let the dimension scale with larger EfficientNet backbone (L2 uses 1376).
Rows 5 and 6 show that using fewer in-batch negatives (50\% and 25\%) in the softmax loss will degrade the performance.
Rows 7-9 study the effect of the temperature parameter in the softmax loss. Compared to the baseline model that learns the temperature parameter (converged to about 1/64), some hand-selected, fixed temperatures could be slightly better. However, we choose to use the learnable temperature as it performs competitively and makes learning easier. We also notice that the temperature usually quickly decrease to only around 1.2x of the converged values in the first 100k steps, and then slowly converges until the end of training.\looseness=-1

\begin{table}[h!]
\vspace{-3mm}
\centering
\small
\caption{Ablation study of key architecture parameters. Baseline model (first row) is trained with embedding dimension 640, using all negatives in the batch, and a learnable softmax temperature.}
\label{tab:ablation_params}
\vspace{1mm}
\resizebox{0.97\linewidth}{!}{%
\begin{tabular}{l|rrr}
\toprule
\multirow{2}{*}{\textbf{Model}} & \multicolumn{2}{c}{\textbf{MSCOCO}} & \multicolumn{1}{c}{\textbf{ImangeNet KNN}} \\
 & I2T~R@1 & T2I~R@1 & R@1 \\
\hline
\rule{-2pt}{10pt}
B5 + BERT-base     & 51.7 & \textbf{37.5} & 64.6\\
[1mm]
~~~~w/ embedding dim=320 & 50.3 & 34.1 & 64.0 \\
~~~~w/ embedding dim=160 & 47.0 & 34.4 & 63.7 \\
~~~~w/ embedding dim=80  & 42.0 & 29.3 & 61.9 \\
[1mm]
~~~~w/ 50\% in-batch negs & 50.2 & 37.0 & 63.8 \\
~~~~w/ 25\% in-batch negs & 48.7 & 35.8 & 63.3 \\
[1mm]
~~~~w/ softmax temp=1/128 & \textbf{52.2} & 36.5 & \textbf{64.8} \\
~~~~w/ softmax temp=1/64  & \textbf{52.2} & 37.3 & \textbf{64.8} \\
~~~~w/ softmax temp=1/32  & 39.6 & 26.9 & 61.2 \\
\bottomrule
\end{tabular}}
\vspace{-3mm}
\end{table}

\subsection{Pre-training Datasets}
It's also important to understand how the model performs when trained on different  datasets with varying size. For this purpose, we train two models: EfficientNet-B7 + BERT-base and EfficientNet-B3 + BERT-mini on three different datasets: full ALIGN training data, 10\% randomly sampled ALIGN training data, and Conceptual Captions (CC-3M, around 3M images). CC-3M is much smaller so we train the model with 1/10 of the default number of steps. All models are trained from scratch. As shown in Table \ref{tab:ablation_data}, a large scale training set is essential to allow scaling up of our models and to achieve better performance. For instance, models trained on ALIGN data clearly outperform those trained on CC-3M data. On CC-3M, B7+BERT-base starts to overfit and performs even worse than B3+BERT-mini. Conversely, a larger model is required to fully utilize the larger dataset -- the smaller B3+BERT-mini almost saturate at 10\% of ALIGN data, while with the larger B7+BERT-base, there is a clear improvement with full ALIGN data.

\begin{table}[h!]
\vspace{-5mm}
    \centering
    \small
    \caption{Ablation study of different training datasets.}
    \label{tab:ablation_data}
    \vspace{1mm}
    \resizebox{0.97\linewidth}{!}{%
    \begin{tabular}{l|r r r}
        \toprule
        \multirow{2}{*}{\textbf{Model + Data}} & \multicolumn{2}{c}{\textbf{MSCOCO}} & \multicolumn{1}{c}{\textbf{ImangeNet KNN}} \\
        & I2T~R@1 & T2I~R@1 & R@1 \\
        \hline
        \rule{-2pt}{8pt}
        B7 + BERT-base & & & \\
        ~~~~ + ALIGN full data & 55.4 & 41.7 & 69.3 \\
        ~~~~ + ALIGN 10\% data & 52.0 & 39.2 & 68.8 \\
        ~~~~ + CC-3M data & 18.9 & 15.5 & 48.7 \\
        [1mm]
        B3 + BERT-mini & & & \\
        ~~~~ + ALIGN full data & 37.4 & 24.5 & 56.5 \\
        ~~~~ + ALIGN 10\% data & 36.7 & 24.4 & 55.8 \\
        ~~~~ + CC-3M data &22.1 & 17.3 & 48.9 \\
        \bottomrule
    \end{tabular}
    }
\vspace{-3mm}
\end{table}

To understand better how data size scaling wins over the increased noise, we further randomly sample 3M, 6M, and 12M ALIGN training data and compare them with the cleaned CC-3M data on B7+BERT-base model. Table \ref{tab:ablation_data_tradeoff} shows that while the ALIGN data performs much worse than CC data with the same size (3M), the model quality trained on 6M and 12M ALIGN data rapidly catches up. Despite being noisy, ALIGN data outperforms Conceptual Captions with only 4x size.

\begin{table}[h!]
\vspace{-5mm}
    \centering
    \small
    \caption{Tradeoff between training data size and quality.}
    \label{tab:ablation_data_tradeoff}
    \vspace{1mm}
    \resizebox{0.97\linewidth}{!}{%
    \begin{tabular}{l|r r r}
        \toprule
        \multirow{2}{*}{\textbf{Model + Data}} & \multicolumn{2}{c}{\textbf{MSCOCO}} & \multicolumn{1}{c}{\textbf{ImangeNet KNN}} \\
        & I2T~R@1 & T2I~R@1 & R@1 \\
        \hline
        \rule{-2pt}{8pt}
        B7 + BERT-base & & & \\
        ~~~~ + ALIGN 12M data & 23.8 & 17.5 & 51.4 \\
        ~~~~ + ALIGN 6M data & 15.8 & 11.9 & 47.9 \\
        ~~~~ + ALIGN 3M data & 8.1 & 6.3 & 41.3 \\
        ~~~~ + CC-3M data & 18.9 & 15.5 & 48.7 \\
        \bottomrule
    \end{tabular}
    }
\vspace{-5mm}
\end{table}


\section{Analysis of Learned Embeddings}

\begin{figure}[!htb]
\vspace{-1mm}
\begin{center}
    \centerline{\includegraphics[width=\linewidth]{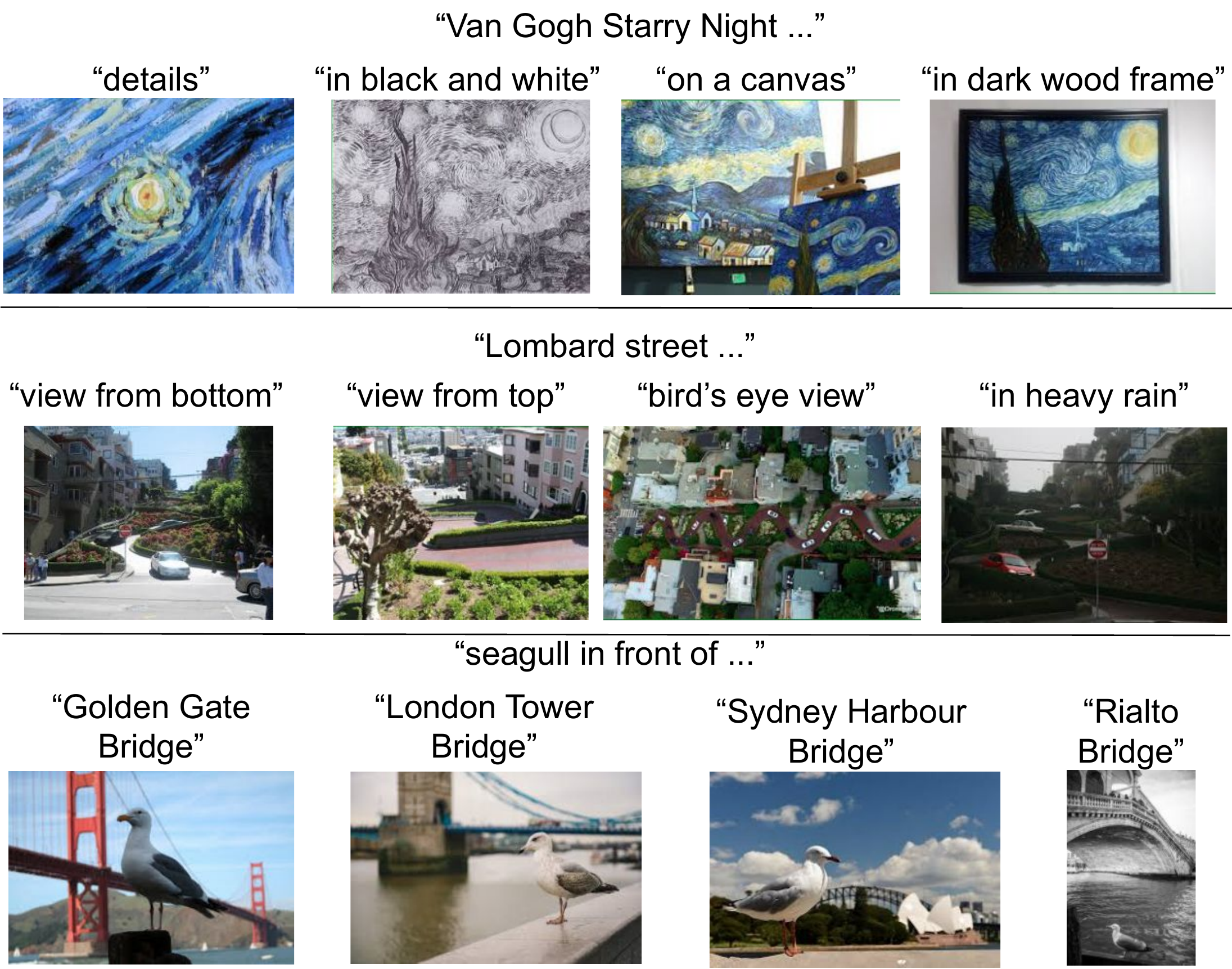}}
    \vskip -0.1in
    \caption{Image retrieval with fine-grained text queries using ALIGN's embeddings.}
    \label{fig:nniq_t2i}
\end{center}
\vspace{-6mm}
\end{figure}

We build a simple image retrieval system to study the behaviors of embeddings trained by ALIGN. For demonstration purposes, we use an index consisting of 160M CC-BY licensed images that are separate from our training set. Figure \ref{fig:nniq_t2i} shows the top 1 text-to-image retrieval results for a handful of text queries not existing in the training data. ALIGN can retrieve precise images given detailed descriptions of a scene, or fine-grained or instance-level concepts like landmarks and artworks. These examples demonstrate that our ALIGN model can align images and texts with similar semantics, and that ALIGN can generalize to novel complex concepts.\looseness=-1

\begin{figure}[h!]
\begin{center}
    \centerline{\includegraphics[width=\linewidth]{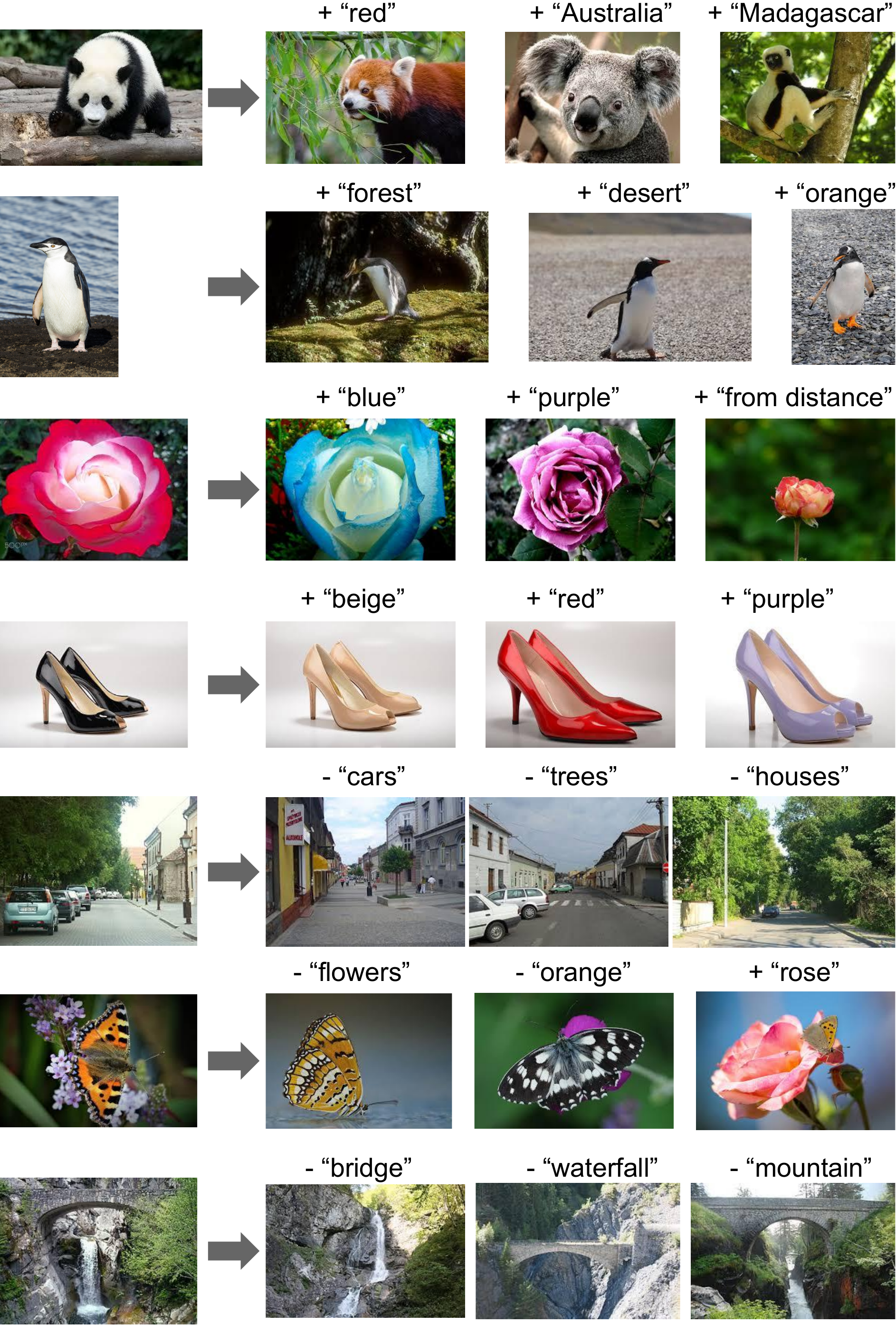}}
    \vskip -0.1in
    \caption{Image retrieval with image$\pm$text queries. We add (or subtract) 
    text query embedding
    to (or from) the 
    image query embedding, and then use the resulting embedding to retrieve relevant images using cosine similarity.}
    \label{fig:nniq_i+t}
\end{center}
\vspace{-10mm}
\end{figure}

Previously word2vec~\cite{mikolov2013efficient, mikolov2013distributed} shows that linear relationships between word vectors emerge as a result of training them to predict adjacent words in sentences and paragraphs. We show that linear relationships between image and text embeddings also emerge in ALIGN. We perform image retrieval using a combined image+text query. Specifically, given a query image and a text string, we add their ALIGN embeddings together and use it to retrieve relevant images.\footnote{We normalize the text and image embeddings before adding them. We also tried various scale factor and found that a scale of 2 for the text embedding and 1 for the image embedding give best results as shown in the figure, although 1:1 also works well.} Figure \ref{fig:nniq_i+t} shows results for a variety of image+text queries. These examples not only demonstrate great compositionality of ALIGN embeddings across vision and language domains, but also show the feasibility of a new paradigm of ``search with multi-modal query" that would otherwise be hard using only text query or image query. For instance, one could now look for the ``Australia" or ``Madagascar" equivalence of pandas, or turn a pair of black shoes into identically-looking shoes with the color of ``beige". Finally, as shown in the last three rows of Figure~\ref{fig:nniq_i+t}, removing objects/attributes from a scene is possible by performing subtraction in the embedding space.

\section{Multilingual ALIGN Model}

One advantage of ALIGN is that the model is trained on noisy web image text data with very simple filters, and none of the filters are language specific.
Given that, we further lift the language constraint of the conceptual caption data processing pipeline to extend the dataset to multilingual (covering 100+ languages) and match its size to the English dataset (1.8B image-text pairs).
A multilingual model ALIGN\textsubscript{mling} is trained using this data.
We created a new mutlilingual wordpiece vocabulary with size 250k to cover all languages. Model training follows the exact English configuration.\looseness=-1

We test the multilingual model on Multi30k, a multilingual image text retrieval dataset extends Flickr30K~\cite{plummer:flickr30k} to German~(de)~\cite{multi30k-en-de}, French~(fr)~\cite{multi30k-fr} and Czech~(cs)~\cite{multi30k-cs}. The dataset consists of 31,783 images with 5 captions per image in English and German and 1 caption per image in French and Czech. The train/dev/test splits are defined in \citet{young-etal-2014-image}.
We evaluate the zero-shot model performance of ALIGN and compare it with M\textsuperscript{3}P~\cite{huang2020m3p} and UC2~\cite{zhou2021uc2}.
The evaluation metric is mean Recall~(mR), which computes the average score of Recall@1, Recall@5 and Recall@10 on image-to-text retrieval and text-to-image retrieval tasks.

Table \ref{tab:multi30k_results} shows that the zero-shot performance of ALIGN\textsubscript{mling} outperforms M\textsuperscript{3}P on all languages by a large margin, with the largest +57.8 absolution mR improvement on fr. The zero-shot performance of ALIGN\textsubscript{mling} is even comparable to the fine-tuned (w/ training splits) M\textsuperscript{3}P and UC2 except on cs. On en, ALIGN\textsubscript{mling} performs slightly worse on its counterpart ALIGN\textsubscript{EN} (trained on EN-only data.)\looseness=-1

\begin{table}[h!]
\begin{center}
\caption{Multimodal retrieval performance on Multi30K dataset. The metric is the mean Recall~(mR).}
\label{tab:multi30k_results}
\begin{small}
\begin{tabular}{l|rrrr}
\toprule
\textbf{Model} & \textbf{en} & \textbf{de} & \textbf{fr} & \textbf{cs} \\
\midrule
\multicolumn{5}{l}{\textit{zero-shot}}\\
~~~~M\textsuperscript{3}P                & 57.9 & 36.8 & 27.1 & 20.4 \\
~~~~\textbf{ALIGN\textsubscript{EN}}     & \textbf{92.2} & - & - & - \\
~~~~\textbf{ALIGN\textsubscript{mling}}  & 90.2 & 84.1 & \textbf{84.9} & 63.2 \\
\hline
\multicolumn{5}{l}{\textit{\rule{-2pt}{8pt} w/ fine-tuning}}\\
~~~~M\textsuperscript{3}P  & 87.7 & 82.7 & 73.9 & 72.2 \\
~~~~UC2  & 88.2 & \textbf{84.5} & 83.9 & \textbf{81.2} \\
\bottomrule
\end{tabular}
\end{small}
\end{center}
\vspace{-7mm}
\end{table}

\section{Conclusion}

We present a simple method of leveraging large-scale noisy image-text data to scale up visual and vision-language representation learning. Our method avoids heavy work on data curation and annotation, and only requires minimal frequency-based cleaning. On this dataset, we train a simple dual-encoder model using a contrastive loss. The resulting model, named ALIGN, is capable of cross-modal retrieval and significantly outperforms SOTA VSE and cross-attention vision-language models. In visual-only downstream tasks, ALIGN is also comparable to or outperforms SOTA models trained with large-scale labeled data.

\section{Social Impacts and Future Work}
While this work shows promising results from a methodology perspective with a simple data collection method, additional analysis of the data and the resulting model is necessary before the use of the model in practice. For instance, considerations should be made towards the potential for the use of harmful text data in alt-texts to reinforce such harms. On the fairness front, data balancing efforts may be required to prevent reinforcing stereotypes from the web data. Additional testing and training around sensitive religious or cultural items should be taken to understand and mitigate the impact from possibly mislabeled data.

Further analysis should also be taken to ensure that the demographic distribution of humans and related cultural items like clothing, food, and art do not cause model performance to be skewed. Analysis and balancing would be required if such models will be used in production. 

Finally, unintended misuse of such models for surveillance or other nefarious purposes should be prohibited.

\section*{Acknowledgements}

This work was done with invaluable help from colleagues from Google. We would like to thank Jan Dlabal and Zhe Li for continuous support in training infrastructure, Simon Kornblith for building the zero-shot \& robustness model evaluation on ImageNet variants, Xiaohua Zhai for help on conducting VTAB evaluation, Mingxing Tan and Max Moroz for suggestions on EfficientNet training, Aleksei Timofeev for the early idea of multimodal query retrieval, Aaron Michelony and Kaushal Patel for their early work on data generation, and Sergey Ioffe, Jason Baldridge and Krishna Srinivasan for the insightful feedback and discussion.





\bibliography{align}

\bibliographystyle{icml2021}

\clearpage
\newpage

\appendix

\section{Remove Near-Duplicate Test Images from Training Data} \label{sec:near-dup}
To detect near-duplicate images, we first train a separate high-quality image embedding model following \cite{wang:sbv3} with a large-scale labeled dataset as in \cite{juan:graphrise}, and then generate 4K clusters via k-means based on all training images of the embedding model. For each query image (from the ALIGN dataset) and index image (from test sets of downstream tasks), we find their top-10 nearest clusters based on the embedding distance. Each image is then assigned to $10 \choose 3$ buckets (all possible combinations of 3 clusters out of 10). For any query-index image pair that falls into the same bucket, we mark it as near-duplicated if their embedding cosine similarity is larger than 0.975. This threshold is trained on a large-scale dataset built with human rated data and synthesized data with random augmentation.

\section{Evaluation on SimLex-999}

The image-text co-training could also help the natural language understanding as shown in \citet{kiros:picturebook}. 
For instance, with language only, it is very hard to learn antonyms.
In order to test this capability of ALIGN model, we also evaluate the word representation from ALIGN model\footnote{As ALIGN uses the wordpiece tokens, one word can be split into multiple pieces. We feed the wordpieces of a word into ALIGN model and use the [CLS] token representation before the project layers as the word embeddings.} on SimLex-999 ~\cite{simlex-999}, which is a task to compare word similarity for 999 word pairs.
We follow \citet{kiros:picturebook} to report the results on 9 sub-tasks each contains a subset of word pairs: \textit{all, adjectives, nouns, verbs, concreteness quartiles (1-4)}, and \textit{hard}.

\begin{table}[h]
\begin{center}
\caption{SimLex-999 results (Spearman’s $\rho$).}
\label{tab:simlex}.
\begin{small}
\begin{tabular}{l|rrr}
\toprule
            & GloVe & Picturebook & \textbf{ALIGN} \\
\midrule
all         & \textbf{40.8} & 37.3 & 39.8 \\
adjs        & \textbf{62.2} & 11.7 & 49.8 \\
nouns       & 42.8 & \textbf{48.2} & 45.9 \\
verbs       & \textbf{19.6} & 17.3 & 16.6 \\
conc-q1     & \textbf{43.3} & 14.4 & 23.9 \\
conc-q2     & 41.6 & 27.5 & \textbf{41.7} \\
conc-q3     & 42.3 & 46.2 & \textbf{47.6} \\
conc-q4     & 40.2 & \textbf{60.7} & 57.8 \\
hard        & 27.2 & 28.8 & \textbf{31.7}\\
\bottomrule
\end{tabular}
\end{small}
\end{center}
\vskip -0.1in
\end{table}

The results are listed in the Table \ref{tab:simlex} compared to Picturebook~\cite{kiros:picturebook} and GloVe~\cite{pennington:glove} embeddings. Overall the learned ALIGN perform better than Picturebook but slightly worse than GloVe embeddings. What is interesting is that the ALIGN word embeddings has a similar trend of Picturebook embeddings, with better performance on \textit{nouns} and \textit{most concrete} categories but worse on \textit{adjs} and \textit{less concrete} categories compared to GloVe embeddings. 
ALIGN word embedding achieves the highest performance on the \textit{hard} category, which similarity is difficult to distinguish from relatedness. This observation confirmed the hypothesis from \citet{kiros:picturebook} that image-based word embeddings are less likely to confuse similarity with relatedness than text learned distributional-based methods.

\end{document}